\documentclass[10pt,twocolumn,letterpaper]{article}

\usepackage{iccv}              
\usepackage{multirow}
\usepackage{multicol}

\definecolor{iccvblue}{rgb}{0.21,0.49,0.74}
\usepackage[pagebackref,breaklinks,colorlinks,allcolors=iccvblue]{hyperref}

\usepackage[accsupp]{axessibility}
\usepackage{hyperref}
\usepackage{url}
\usepackage{xcolor}
\usepackage{color, colortbl}
\usepackage{graphicx}
\usepackage{multirow}
\usepackage{threeparttable}
\usepackage{bbding}
\usepackage{tabularx}
\usepackage{makecell}
\usepackage{subcaption}
\usepackage{wrapfig}
\usepackage{longtable}

\usepackage{tabularx}
\usepackage{algorithm}
\usepackage{algorithmic}
\usepackage{listings}
\usepackage{pifont}

\def\paperID{8020} 
\def\confName{ICCV}
\def\confYear{2025}
\definecolor{mydarkgreen}{rgb}{0.2,0.7,0.2}
\title{LVAgent: Long Video Understanding by \\  Multi-Round Dynamical Collaboration of MLLM Agents}

\author{Boyu Chen\textsuperscript{\rm1,\rm2 \thanks{Equal contribution.}}, Zhengrong Yue\textsuperscript{\rm5,\rm6 \footnotemark[1]}, Siran Chen\textsuperscript{\rm1,\rm2 \footnotemark[1]}, Zikang Wang\textsuperscript{\rm5,\rm6 \footnotemark[1]}, Yang Liu\textsuperscript{\rm3,\rm4,\rm5}, Peng Li\textsuperscript{\rm3,\rm5 \thanks{Equal corresponding author.}}, Yali Wang\textsuperscript{\rm1,\rm5 \footnotemark[2]}\\
\textsuperscript{\rm1}Shenzhen Key Lab of Computer Vision and Pattern Recognition, Shenzhen Institutes of \\ Advanced Technology, Chinese Academy of Sciences\\
\textsuperscript{\rm2}School of ArtificialIntelligence, University of Chinese Academy of Sciences \\
\textsuperscript{\rm3}Institute for AI Industry Research (AIR), Tsinghua University, Beijing, China\\
\textsuperscript{\rm4}Dept. of Comp. Sci. \& Tech., Institute for AI, Tsinghua University, Beijing, China\\
\textsuperscript{\rm5}Shanghai Artificial Intelligence Laboratory \\
\textsuperscript{\rm6}Shanghai Jiao Tong University \\
}

\begin{document}
\maketitle
\begin{abstract}
Existing MLLMs encounter significant challenges in modeling the temporal context within long videos. 
Currently, mainstream Agent-based methods use external tools to assist a single MLLM in answering long video questions. Despite such tool-based support, a solitary MLLM still offers only a partial understanding of long videos, resulting in limited performance.
In order to better address long video tasks, we introduce \textbf{LVAgent}, the first framework enabling multi-round dynamic collaboration of MLLM agents in long video understanding.
Our method consists of four key steps: 
1) \textbf{Selection:} We pre-select appropriate agents from the model library to form optimal agent teams based on different tasks.
2) \textbf{Perception:} 
We design an effective retrieval scheme for long videos, improving the coverage of critical temporal segments while maintaining computational efficiency‌.
3) \textbf{Action:} Agents answer long video questions and exchange reasons.
4) \textbf{Reflection:} We evaluate each agent's performance in each round of discussion and optimize the agent team for dynamic collaboration.
The agents iteratively refine their answers by multi-round dynamical collaboration of MLLM agents.
LVAgent is the first agent system method that outperforms all closed-source models (like GPT-4o) and open-source models (like InternVL-2.5 and Qwen2-VL) in the long video understanding tasks.  
Our LVAgent achieves an accuracy of 80\% on four mainstream long video understanding tasks. Notably, LVAgent improves accuracy by 13.3\% on LongVideoBench. Code is available at \url{https://github.com/64327069/LVAgent}.

\end{abstract}
    
\section{Introduction}
\label{sec:intro}
\begin{figure}[h]
\vspace{-1.5em}
    \centering
    \includegraphics[width=0.45\textwidth]{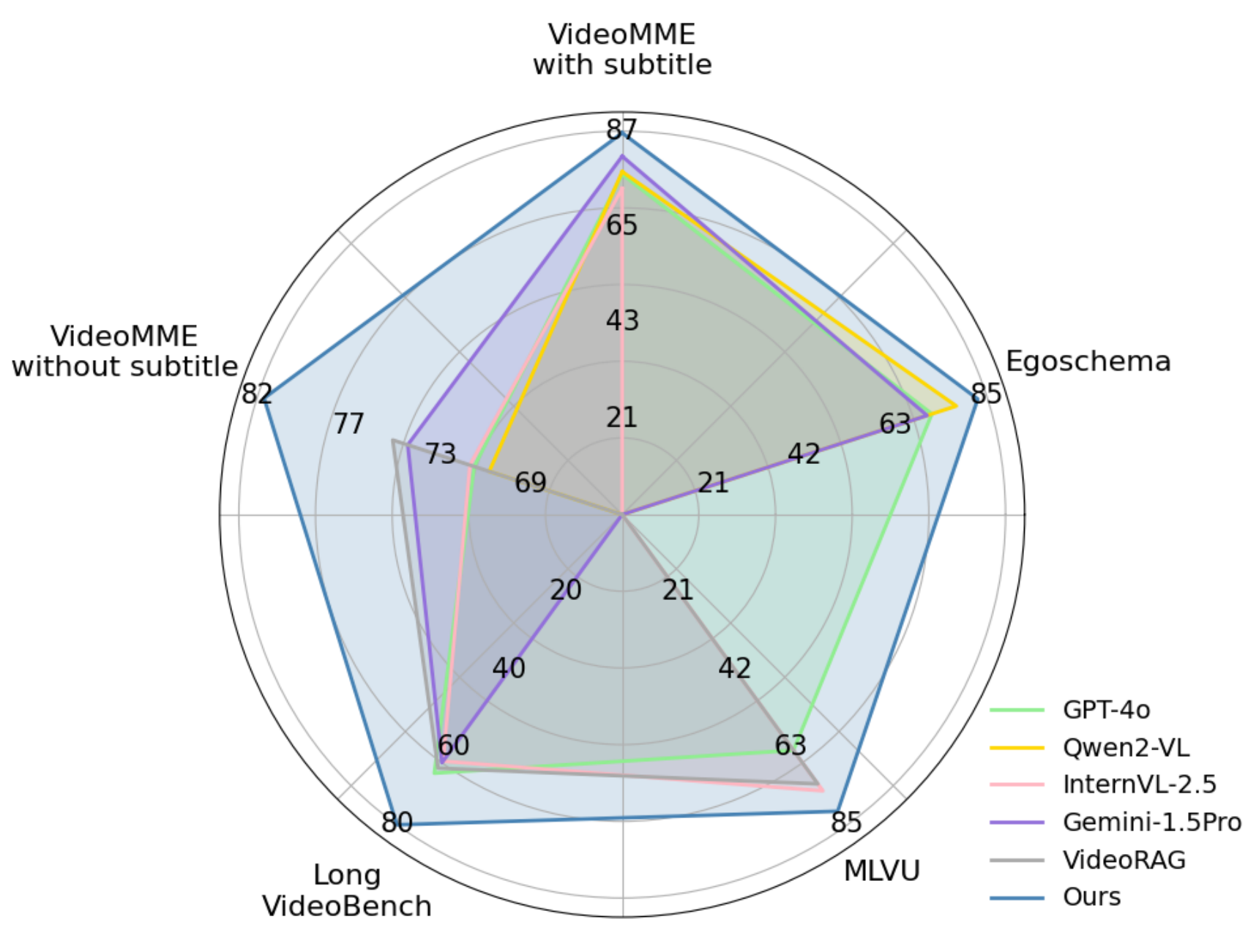}
    \vspace{-0.5em}
    \caption{
        \textbf{Comparsion with SOTA.} LVAgent outperforms all closed-source models (including GPT-4o) and open-source models (including Agent-based methods) in long video tasks.
    }
    \label{fig:dataset}
    \vspace{-1.5em}
\end{figure}

Long videos are crucial in Internet data and important in fields like healthcare, education, and entertainment~\cite{videoinedu,videoinmed,videoinent}. 
In the current data-driven era, correctly and efficiently understanding long videos holds great significance. 
Despite being pretrained on large multi-modal datasets, Multi-modal Large Language Models (MLLMs)~\cite{gpt4o,gemini,qwen2vl,liu2024nvila,longvila,timesuite,videoxl,videollama,moviechat,chen2022low,chen2025g} struggle with modeling the long-term temporal context of long videos (from minutes to hours long). 
Feeding numerous video frames (e.g., at 1 frame/s) directly into a single MLLM incurs high computational costs and much redundant information, yielding poor performance. 
Taking Figure~\ref{fig:visualization} as an example, we input the whole video into Gemini 1.5Pro~\cite{gemini15}, GPT-4o~\cite{gpt4o}, Qwen2-VL~\cite{qwen2vl} and InternVL-2.5~\cite{internvl2.5}, and all of these models fail to provide the correct answer due to the redundant confusing information of the video, as shown with colored text.

Agent-based approaches~\cite{videoagentmemory,videoagentrepeat,videorag,videotree, vca,chen2025vragent, chen2025videochat, yue2025uniflow, chen2025top} hold great potential in long video understanding, because agents can simplify difficult problems and autonomously invoke a variety of tools to extract key information from long videos to assist in long video understanding. 
However, there is still significant room for improvement in current agent-based methods for long video tasks.
Some methods use CLIP~\cite{clip} to retrieve key frames for long video understanding~\cite{videoagentrepeat,vca,zhao2024longagent,xie2024openagents}. However, CLIP struggles to retrieve long-term temporal information and its pretrained data has a domain gap with long videos~\cite{videoclipxl,longclip}. 
Some approaches use external tools (e.g., memory bank~\cite{videoagentmemory,mallm}, RAG~\cite{searchlvlm, chen2024sharegpt4video,videorag}) to aid MLLMs in long video understanding. 
Even with the support of tools, relying on a single  MLLM to answer questions about long-form videos offers only a partial comprehension of the video content, leading to limited performance.
In conclusion, there are two main challenges in long video understanding.
The first challenge is how to better retrieve the video clip with key information according to the query.
The second challenge is how to use multi-agent collaboration to better understand long videos.

\begin{figure}
    \centering

    \includegraphics[width=\linewidth]{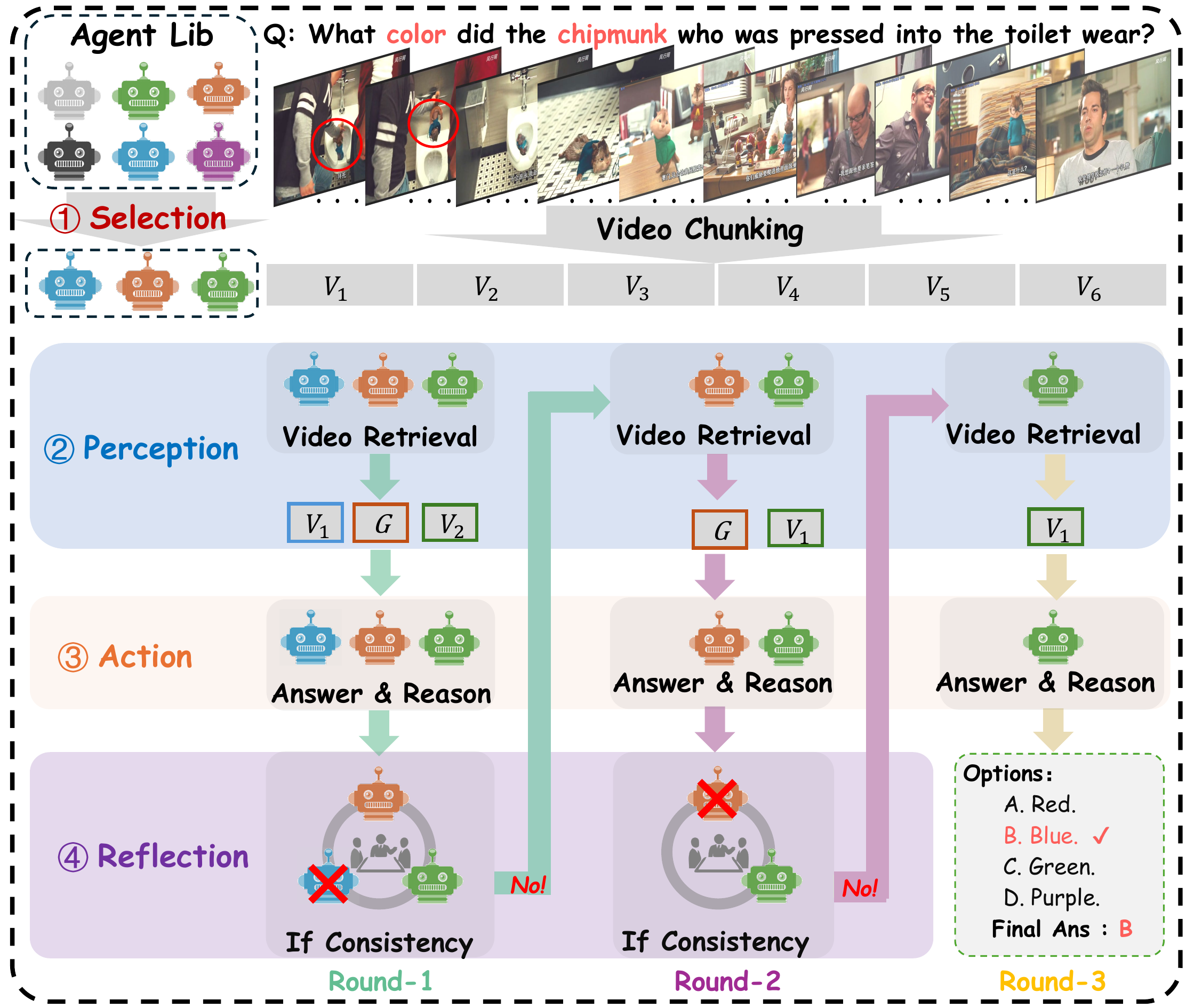}
    \vspace{-1.em}
    
    \caption{\textbf{The Overview of LVAgent.} LVAgent features four key processes: Selection, Perception, Action, and Reflection.
    In the perception process, we retrieve the video chunk related to the question. $G$ means global sampling of the long video.
    }
    \label{fig:enter-label}
    \vspace{-1.5em}
\end{figure}

To address these challenges, we put forward a four-step dynamic collaborative paradigm for agents, namely Selection, Perception, Action, and Reflection.
(1) \textbf{Selection}. We pre-select the optimal agent team from an agent library housing mainstream MLLMs like Qwen2-VL~\cite{qwen2vl} and InternVL-2.5~\cite{internvl2.5}. With selection process, only the most suitable models are involved in the subsequent processes, thus saving computational resources and enhancing efficiency.
(2) \textbf{Perception}. Regarding the challenge of CLIP in retrieving long video temporal info, we propose an adaptive frame extraction method and design an effective retrieval approach for long videos.
This pipeline is designed to enhance the coverage of crucial temporal chunks without sacrificing computational efficiency. 
(3) \textbf{Action}. After retrieving the query-related video clips, agents answer long video questions and provide the reasons.
(4) \textbf{Reflection}. In order to enable better collaboration among multiple agents for a better understanding of long videos, 
we introduce a reflection process to evaluate each agent's performance in every discussion round. Based on performance, agents are scored and filtered, enabling dynamic collaboration for each query. 
This reflection process helps facilitate better collaboration, rule out wrong inferences, and reach a consensus. 
Throughout these four processes, agents refine answers by discussing, retrieving key video chunks, and integrating information with other pre-selected agents.
We validate the effectiveness on four long video understanding datasets: EgoSchema~\cite{egoschema}, VideoMME~\cite{videomme}, MLVU~\cite{MLVU}, LongVideoBench~\cite{wu2024longvideobench}. 
Experiments show that LVAgent achieves an accuracy of over 80\% on mainstream long video tasks. Our contributions can be summarized as threefold:

\begin{figure}
    \centering
    \includegraphics[width=0.95\linewidth]{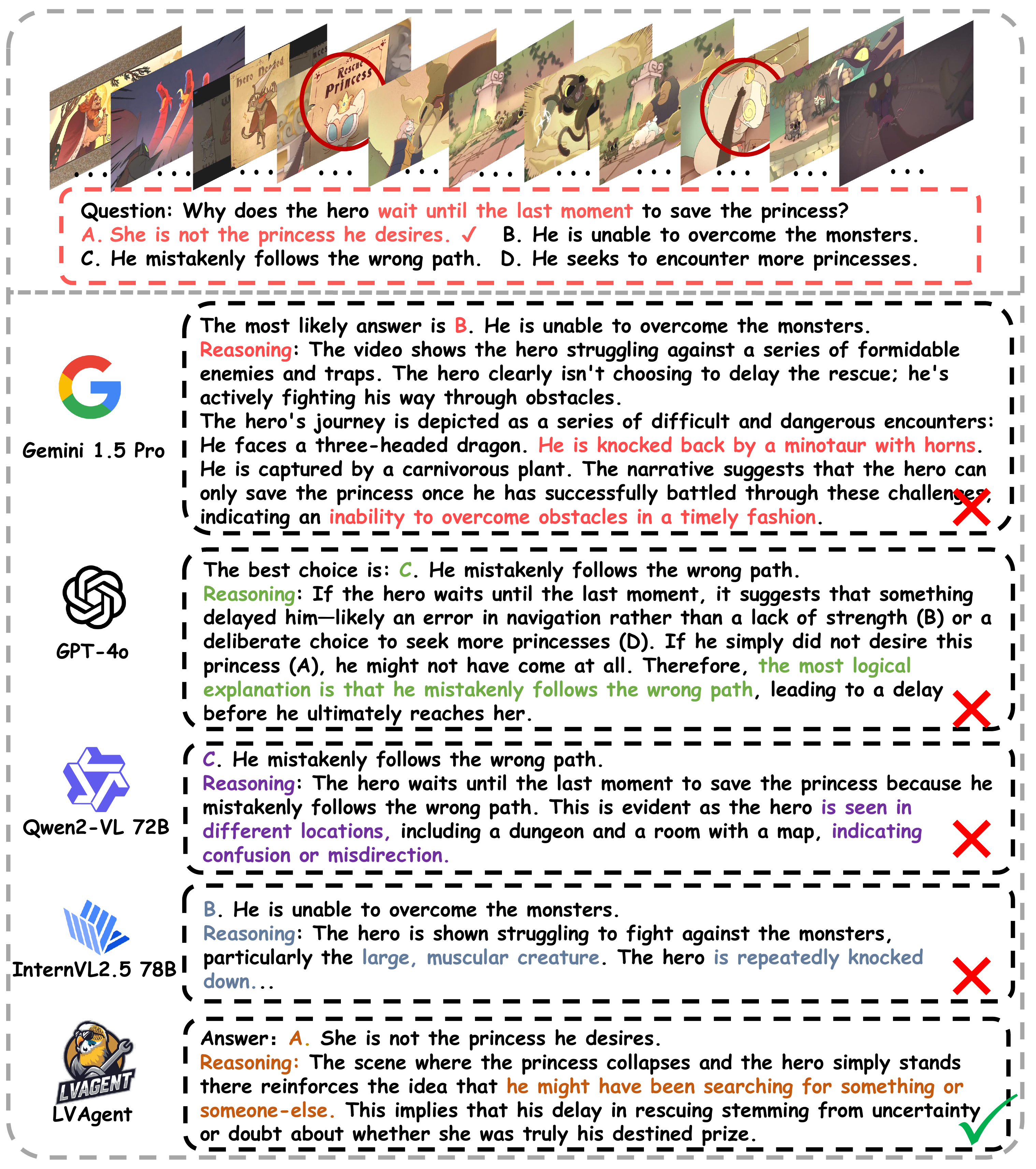}
    \vspace{-0.7em}
    
    \caption{\textbf{Visualization of LVAgent.} The visualized results of different models’ responses to a specific question and the reasons for their answers, with key parts highlighted in different colors.
    }
    \label{fig:visualization}
    \vspace{-1.5em}
\end{figure}

\begin{figure*}[h]
    \centering
    \includegraphics[width=0.98\textwidth]{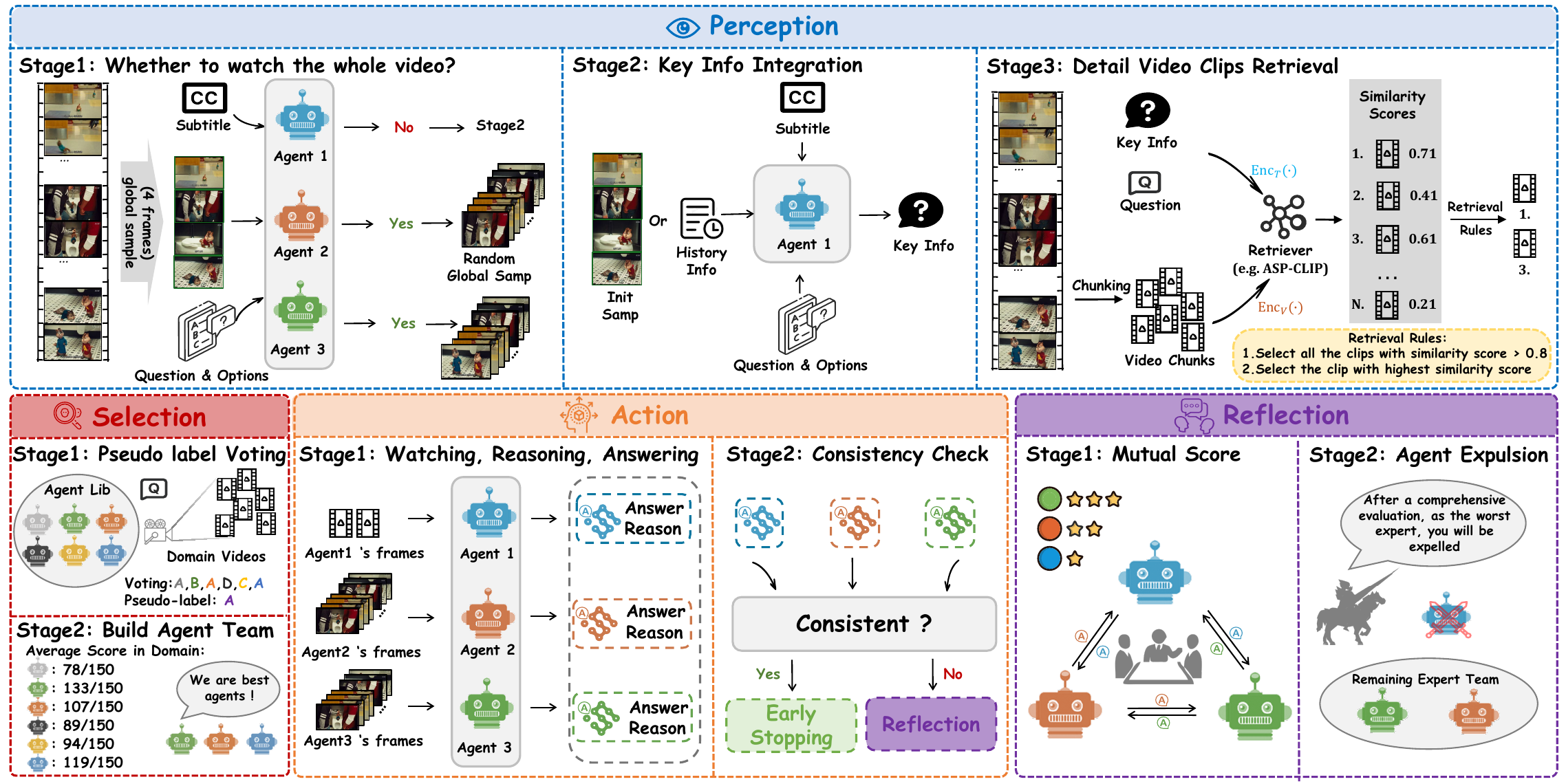}
    \vspace{-1em}
    \caption{\textbf{LVAgent Framework.} This diagram illustrates four key steps of our LVAgent collaboration framework: Selection, Perception, Action, and Reflection. }
    
    
    \vspace{-1.5em}
    
    \label{fig:method}
\end{figure*}

\begin{itemize}

    \item First, we propose LVAgent, a long video multi-agent collaboration pipeline for long video understanding.

    \item ‌Second, LVAgent is a novel multi-round and multi-step dynamical collaboration pipeline of MLLM Agents. LVAgent enables multi MLLM agents to process and integrate long-term temporal context dynamically for more comprehensive and robust long video understanding.
    
    \item Finally, empirical results demonstrate the superior accuracy of our LVAgent. LVAgent achieves an accuracy exceeding 80\% across four mainstream long video understanding tasks. It is the first multi-round dynamic multi-agent collaboration pipeline that outperforms all closed-source models (including GPT) and open-source models (including InternVL-2.5 and Qwen2-VL) in the long video understanding tasks. 
    Notably, on the LongVideoBench dataset, LVAgent improves accuracy by up to 13.3\% compared with SOTA.

\end{itemize}

\section{Related Work}

\paragraph{Long Video MLLMs.}
Recent advancements in multimodal large language models (MLLMs), such as~\cite{qwen2vl, videochat, llavavideo, mvbench, llavaonevision, timesuite, longvu}, have significantly advanced video understanding, particularly in temporal reasoning. However, processing long videos remains a critical challenge due to the inherent redundancies in temporal sequences, which result in an overwhelming number of visual tokens. 
To address this, current approaches primarily focus on two strategies: (1) extending the context length of LLMs~\cite{lwm, gemini, longllava, longvila} and (2) compressing video tokens~\cite{moviechat, videollama, longvlm, llamavid, videoccam, videoxl, timesuite}.
For context extension, methods like LongVA~\cite{longva} fine-tune language models to expand their context windows, successfully generalizing short-video understanding to longer videos. 
However, context extension often overlooks redundant information and computational overhead caused by the increase in the number of frames during inference. 
On the other hand, token compression techniques, such as MovieChat~\cite{moviechat}, reduce redundancy by merging similar visual tokens. While this improves efficiency, it often sacrifices fine-grained details, leading to suboptimal performance in complex scenarios.

\label{method}
\begin{figure*}
    \centering
    \includegraphics[width=0.9\linewidth]{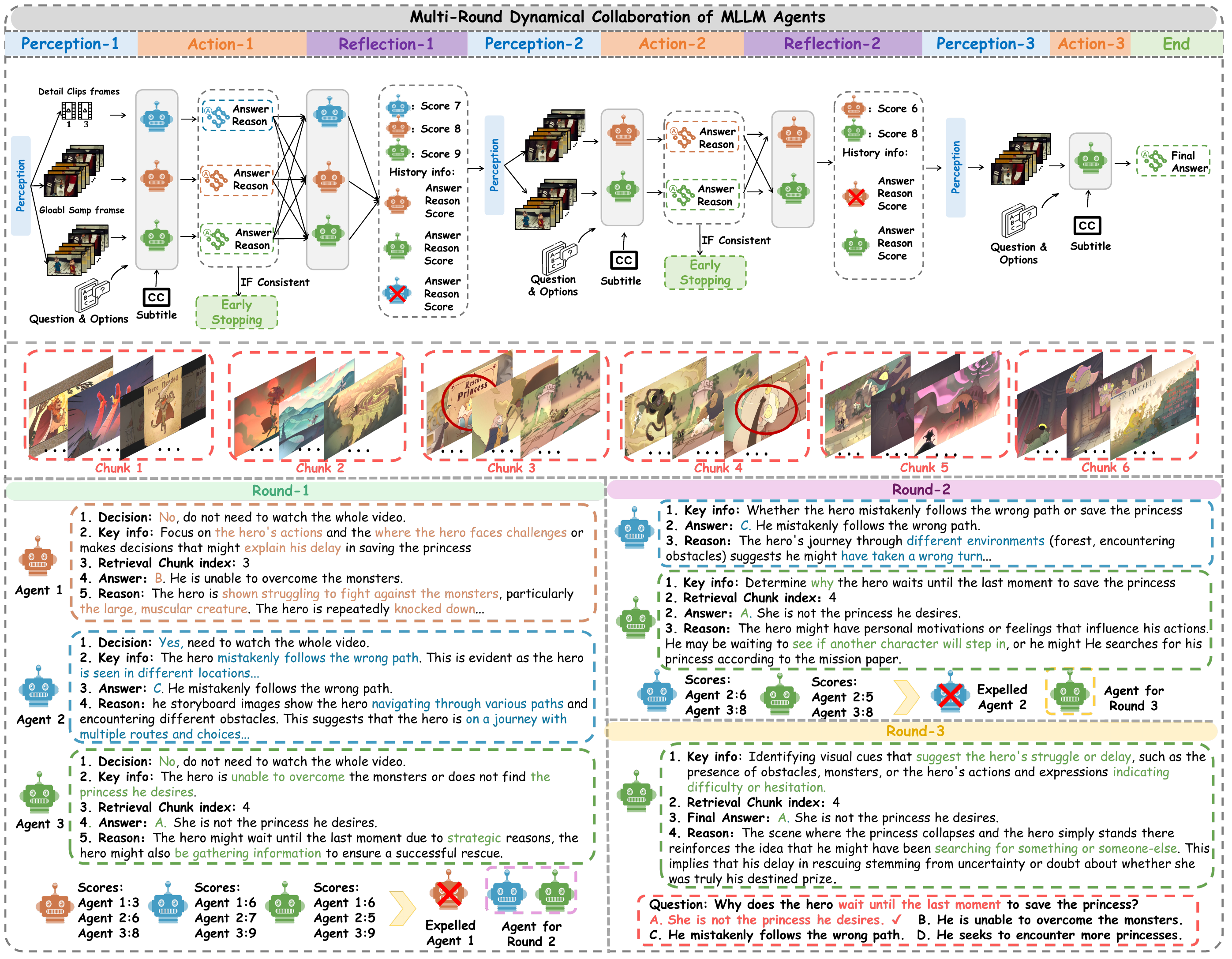}

    \caption{\textbf{The Multi-Round Dynamic Collaboration of LVAgent.} The upper part is our collaboration framework, where 3 rounds of discussion were conducted for this question. The lower part presents the results and discussion details of each round. The detailed prompt can be found in supplementary document.
    }
    \label{fig:discuss}
    \vspace{-1em}
\end{figure*}

\paragraph{Long Video Agent.}
Agent-based methods~\cite{li2023camel, debate1-mit, multi-persona, blender, reflexion, autogen, pca} have the ability to decompose complex tasks.
Currently, there are mainly three paradigms for agents in solving long video tasks. 
First, some agent methods employ retrieval methods such as CLIP~\cite{clip} to identify the video regions most relevant to the question~\cite{zhao2024longagent,xie2024openagents, videoagentrepeat,chen2025super}. However, CLIP-based approach suffers from a domain gap when dealing with long videos.
Second, the Retrieval Augmented Generation (RAG) method~\cite{videorag} leverages external tools for information extraction, such as memory banks~\cite{videoagentmemory, mallm} and search engines~\cite{searchlvlm,chen2024sharegpt4video}, for a more comprehensive understanding.
However, the accuracy of the RAG method is also limited.
Third, the Chain-of-thought (CoT) method is adopted to enhance the model's reasoning ability to answer questions correctly~\cite{videotree,videoofthought}. 
While agents show promise for long video tasks, current agent-based methods rely only on a single MLLM for such questions, lacking a dynamic framework for multiple MLLMs to answer, discuss, and collaborate. This ineffective collaboration results in an incomplete and partial understanding of long videos.
There is still much room for improvement in accuracy. Specifically, none of the existing agent paradigms has outperformed the best Multimodal Large Language Models (MLLMs)~\cite{qwen2vl,gpt4o,llavavideo,gemini15} on the long video understanding tasks~\cite{egoschema,videomme,zhou2024mlvu,wu2024longvideobench}.
Fortunately, multi-round dynamical collaboration pipeline~\cite{liu2024dynamic} demonstrated markable performance in NLP tasks. Inspired by this, we introduce our multi-round dynamical collaboration of MLLM agents for long video understanding.

\section{Method}

In the following section, we will comprehensively introduce our proposed LVAgent paradigm, with four key processes, namely Selection, Perception, Action, and Reflection, in a detailed and sequential manner.

\subsection{Selection}  
\label{subsec: selection}
To answer the question about long videos, we leverage the popular MLLMs to establish an \textbf{Agent Library}, 
which encompasses currently popular MLLMs, i.e. Qwen2-VL~\cite{qwen2vl}, InternVL-2.5~\cite{internvl2.5}, LongVU~\cite{longvu}, and LLaVA-Video (LV)~\cite{llavavideo}. 
In multi-agent systems, agents often vary in performance across tasks and domains, with underperforming agents disrupting collaboration and increasing computational costs. To address this, we propose an agent pre-selection method based on pseudo-label voting to efficiently identify the optimal agent team for a task as shown in Figure~\ref{fig:method}.

To quickly assess the performance of each agent $A_i (i=1,2,..,N)$,  where $N$ is the total number of agents in the agent library,
we randomly sample a subset $S$ of 150 videos from the task dataset $D$, \textbf{without labels}.
For each video case $V\in S$ which comes with a question $Q$, the video subtitle $T$, and options $O$, we adopt Perception and Action process to obtain the answer set $S_{ans}=\{ans_0, ans_1,...,ans_n\}$ of video $V$.
The Perception and Action processes will be introduced in detail later.
The answer that appears in $S_{ans}$ most frequently in this set is used as the pseudo label $L_v$ of the video $V$. We use accuracy to measure the performance of each agent $A_i$ with pseudo label, as expressed by the formula:
\begin{equation}
    Acc\left(A_i\right)=\frac{1}{|S|} \sum_{V \in S}\left[A_i(Q,T,O,V_{samp})=L_V\right]
\label{eq:acc}
\end{equation}
The top three agents with the highest accuracy of pseudo labels are selected to form the optimal team. This agent pre-selection process ensures that only the most competent agents advance to subsequent stages, thereby maximizing the efficacy of the collaborative reasoning framework.

\subsection{Perception} 
To reduce the interference of redundant information, retrieve the key regions of the video, and answer questions more effectively, we propose a novel three stage perception pipeline as shown in Figure~\ref{fig:method}.

In stage 1, the agents are enabled to autonomously determine whether it is necessary to watch the entire video before answering the question.
We first randomly sample 4 frames $\hat{v}$ to give each agent $A_i$ a rough understanding of the video $V$.
Then, based on $\hat{v}, Q, T, O$, $A_i$ determines whether to watch the entire video:
\begin{equation}
    D_i = A_i(\hat{v},Q, T, O)
\label{eq:watch}
\end{equation}
If the model decides to watch the entire video (i.e. $D_i$ is ``Yes''), 16 frames are extracted through global sampling.
Conversely, if the model considers it unnecessary to watch the entire video, then it moves on to stage 2.

In stage 2, we summarize the key information for answering the question based on the rough visual information $\hat{v}$, the question $Q$, and the options $O$. In the first round of collaboration, the agents generate the key information of the video, which is formulated as:
\begin{equation}
    K_i = A_i(\hat{v},Q, T, O)
\label{eq:watch2}
\end{equation}
After the first round of collaboration, the agents will generate the history information $H_{info}$.
 $H_{info}$ consists of each agent's answer, reason, score, and whether it needs to be removed, as shown in Figure~\ref{fig:discuss}
The key information for next round's Perception process is formulated as:
\begin{equation}
    K_i = A_i(H_{info},Q, T, O)
\label{eq:info}
\end{equation}
$K_i$ enables the retrieval model to conduct retrievals. It is important to note that, owing to the disparities among individual agents, the key information $K_i$ and the determination $D_i$ of whether to view the entire video can be different.

In stage 3, we perform detail video chunks retrieval.
Through ablation experiments, we divide the video into six equal chunks $\{chk_1, chk_2,...,chk_6 \}$. For each chunk, we randomly sample 16 frames to form a frame set $S_{frame} = {f_1,f_2,...,f_6}$. Then, we use ASP-CLIP model\cite{asp} finetuned on our collected LongVR dataset to calculate the CLIP score for each video chunk.
\begin{equation}
    Sim(chk_i) = ASP(f_i, K_i) + ASP(f_i, Q)
\label{eq:samp}
\end{equation}
If the overall CLIP score of a chunk is greater than 0.8, the sampled frames $f_i$ in this chunk $chk_i$ are selected. If no chunk has a score greater than 0.8, 
we select the sampled frames in the chunk with the highest CLIP score.

This approach effectively reduces the difficulty of retrieval by transforming the long video retrieval problem into a simple multi-classification problem with six categories (chunks). Ablation experiments show that our proposed  perception process improves the accuracy of the model in handling long videos.

\begin{figure}
    \centering
    \includegraphics[width=0.95\linewidth]{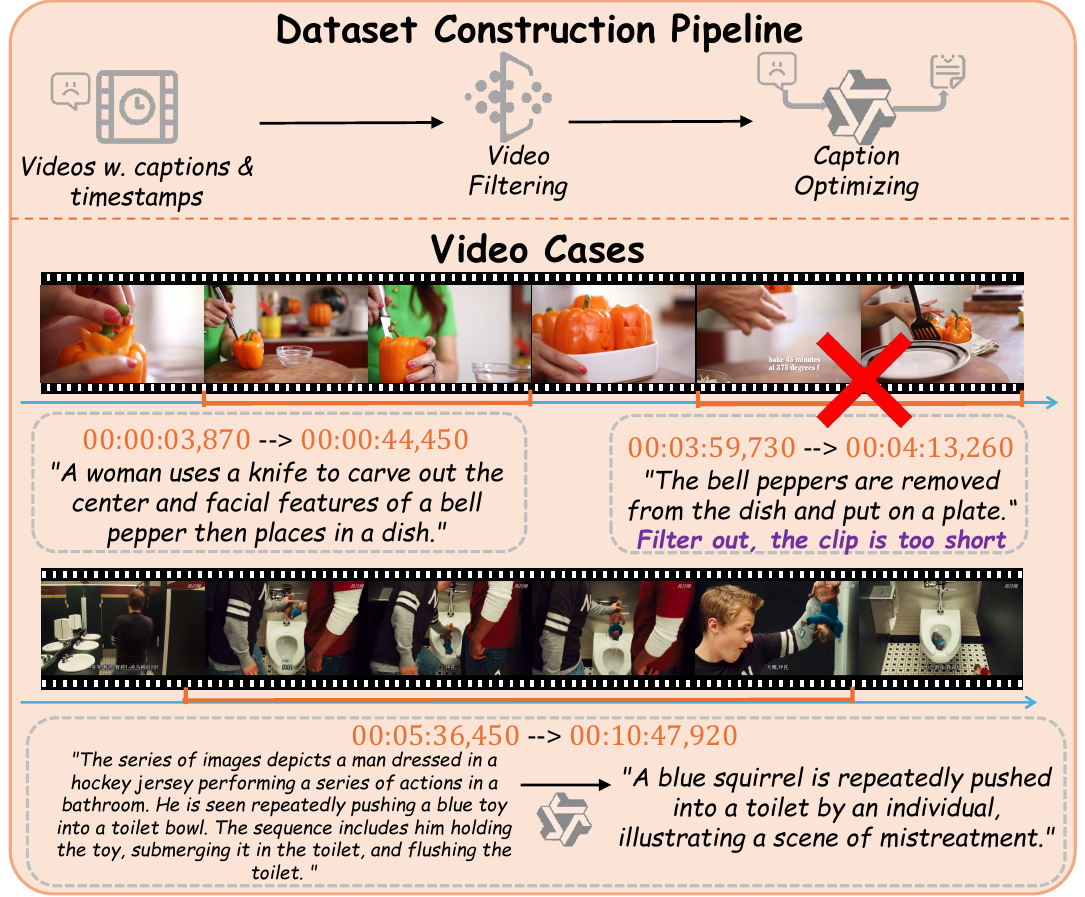}
    \vspace{-0.7em}
    
    \caption{\textbf{LongVR Data Pipeline.} The construction of LongVR for long video chunks retrieval in perception process.}
    
    \label{fig:longvr}
    \vspace{-1.3em}
\end{figure}

\paragraph{LongVR dataset for Long Video Retrieval.}
To enhance the retrieval ability of ASP-CLIP to better assist with long video tasks,
we compile a diverse dataset for fine-tuning the ASP-CLIP model, 
consisting of 82k video clips from five datasets, as shown in Table~\ref{tab:videoclip_dataset} in the appendix. 
This collection covers a wide range of scenes, themes, and video quality, providing a robust foundation for training a video retrieval model.
We extract long videos from the datasets of ActivityNet~\cite{activitynet}, OpenVid-1M~\cite{openvid}, ViTT~\cite{vitt}, MovieChat-Caption~\cite{moviechat}, and Youcook2~\cite{youcook2} for the retrieval task, as depicted in Figure \ref{fig:longvr}. 
First, we crop the videos clips with captions.
If a video has been pre-annotated into multiple clips, we take these clips along with their captions as components of our dataset.
Second,
to better suit the long video scenario, we filter out video clips shorter than 5 seconds or longer than 12 minutes.
Only clips of appropriate length and their captions are retained.
Finally,
considering that the context length of CLIP is limited, to prevent information truncation in the clip-text encoder due to overly long captions, we use Qwen2-VL~\cite{qwen2vl} for caption optimization. 
Additionally, if the number of tokens in a caption is less than 20, we also filter out this data, as such short captions are insufficient to summarize the content of a relatively long video. 
After manually screening the captions, the resulting long video dataset has an average video duration of 145.6 seconds. The captions in the dataset have an average length of 71 tokens.
In total, we assemble a long video retrieval dataset with 82K samples. The details of dataset constructions and data recipes can be found in the supplementary document.
We finetune ASP-CLIP~\cite{asp} on LongVR and achieved significant improvements.

\begin{table*}[t]
\centering
\resizebox{0.9\textwidth}{!}{

\begin{tabular}{lccccccc}
\toprule
\small
\multirow{2}{*}{\textbf{Model}}  & \multirow{2}{*}{\textbf{Egoschema}} & \multirow{2}{*}{\textbf{LongVideoBench}} & \multirow{2}{*}{\textbf{MLVU }} & \multicolumn{4}{c}{\textbf{Video-MME}} \\
\cmidrule(lr){5-8}
&  &  &  \textbf{(M-avg)} & \textbf{Short} & \textbf{Medium} & \textbf{Long} & \textbf{Average}  \\
\midrule
GPT-4o~\cite{gpt4o}  &72.2 & \textcolor{blue}{66.7} & 64.6 & 80.0 / 82.8 & 70.3 / 76.6 & 65.3 / 72.1 & 71.9 / 77.2 \\
Gemini 1.5Pro~\cite{gemini15}  & 71.1 & 64.0 & - & 81.7 / \textcolor{blue}{84.5} & \textcolor{blue}{74.3} / \textcolor{blue}{81.0} & 67.4 / \textcolor{blue}{77.4} & 75.0 / \textcolor{blue}{81.3}\\

\midrule
ShareGPT4Video-8B~\cite{chen2024sharegpt4video}  &- & 39.7 & 46.4 & 48.3 / 53.6 & 36.3 / 39.3 & 35.0 / 37.9 & 39.9 / 43.6 \\
VideoChat2-7B~\cite{mvbench} & 56.7 & 39.3 & 47.9 & 48.3 / 52.8 & 37.0 / 39.4 & 33.2 / 39.2 & 39.5 / 43.8 \\
InternVL-2.5-8B~\cite{internvl2.5}& - & 60.0 & 68.9 & - & - & - & 64.2 / 66.9 \\
Qwen2-VL-7B~\cite{qwen2vl} & 66.7 & 55.6&-&-&-&-&63.3 / 69.0\\
LongVA-7B~\cite{longva} & -& 51.3 & 58.8 & 61.1 / 61.6 & 50.4 / 53.6 & 46.2 / 47.6 & 52.6 / 54.3 \\
LongVU-7B~\cite{longvu} & 67.6 & - &  65.4 & - & - &  - / 59.5 & - / 60.6 \\
LLaVA-Video-7B~\cite{llavavideo} & 57.3 & 58.2 & 70.8 & - & - & - & 63.3 / 69.7 \\
\midrule
Aria-28B~\cite{aria} & - & 64.2 & 72.3& 67.9 / 78.3 & 67.0 / 71.7 & 58.8 / 66.3 & 67.6 / 72.1 \\
Oryx-34B~\cite{liu2024oryx} & - & - & 70.6 & 77.3 / 80.6& 65.3 / 74.3 & 59.3 / 69.9 & 67.3 / 74.9 \\

Qwen2-VL-72B~\cite{qwen2vl} & \textcolor{blue}{77.9} & - & - & 80.1 / 82.2 & 71.3 / 76.8 & 62.2 / 74.3 & 71.2 / 77.8 \\

LLAVA-Video-72B~\cite{llavavideo} & 65.6 & 64.9 & - & 81.4 / 82.8 & 68.9 / 75.6 & 61.5 / 72.5 & 70.6 / 76.9 \\

InternVL-2.5-78B~\cite{internvl2.5} & - & 63.6 & \textcolor{blue}{75.7}  & \textcolor{blue}{82.8} / 83.2 & 70.9 / 74.1 & 62.6 / 64.8 & 72.1 / 74.0 \\

LLaVA-OneVision-72B~\cite{llavaonevision} & 62.0 & 63.2 & 68.0 & 76.7 / 79.3 & 62.2 / 66.9 & 60.0 / 62.4 & 66.3 / 69.6 \\
VideoLLaMA-2-72B~\cite{videollama2} & 63.9 & - & 45.6 & 69.8 / 72.0 & 59.9 / 63.0 & 57.6 / 59.0 & 62.4 / 64.7 \\
\midrule
VideoAgent~\cite{videoagentrepeat} &54.1 & -& -& - & - & - & - \\
VideoTree~\cite{videotree} & 61.1& - & -& - & - & - & - \\
VCA~\cite{vca}  & - & 41.3 & -& - & - & - & - \\
DrVideo~\cite{drvideo}   & 61.0  & -	 & -& - & - & - & 51.7/71.7\\
VideoRAG-72B~\cite{videoragvideofeat}   & - & 65.4 & 73.8 & 81.1 / - & 72.9 / - & \textcolor{blue}{73.1} / - & \textcolor{blue}{75.7} / - \\
\midrule

\multirow{2}{*}{\textbf{LVAgent}~\footnotesize{(\textbf{\texttt{ours}})}} & \textbf{82.9} & \textbf{80.0} & \textbf{83.9}  & \textbf{88.9} / \textbf{90.7}& \textbf{82.0} / \textbf{87.6} & \textbf{74.3} / \textbf{81.7} & \textbf{81.7} / \textbf{86.6}\\

   & \textcolor{mydarkgreen}{\textbf{+5.0\%}}  &\textcolor{mydarkgreen}{\textbf{ +13.3\%}} &\textcolor{mydarkgreen}{\textbf{ +8.2\%}} &\textcolor{mydarkgreen}{\textbf{+6.1\% / +6.2\%}} & \textcolor{mydarkgreen}{\textbf{+7.7\% / +6.6\%}} & \textcolor{mydarkgreen}{\textbf{+1.2\% / +4.3\%}} & \textcolor{mydarkgreen}{\textbf{+6.0\% /+5.3\%}} \\

\bottomrule
\end{tabular}
}
\vspace{-0.6em}

\caption{\textbf{Comparison on Mainstream Long Video Understanding Tasks.} The Video-MME results are presented in the format ``w/o subs / w/ subs". We are the first agent-based method to achieve over 80\% accuracy on each long video understanding dataset}
\vspace{-1.2em}
\label{tab:results}
\end{table*}

\subsection{Action} 
\label{Action}

In the Action process shown in Figure~\ref{fig:method}, we first obtain the answers and the reasons based on the extracted frames on the Perception process: 
\begin{equation}
    ans_i, R_i = A_i(V_{samp},Q,T,O)
\label{eq:ans}
\end{equation}
For all agents within the seleted agent team, we acquire a set of answers $S_{ans}=\{ans_0, ans_1,...,ans_n\}$ along with a corresponding set of reasons $S_R = \{R_0, R_1,...,R_n\}$ that each agent used to arrive at its respective answer.
In stage 2 of the Action process, we conduct a consistency check.
If the proportion of a certain answer given by the agents exceeds half of the total number of agents, the early stopping mechanism will be triggered, and this answer will be selected as the final answer. If the agents fail to reach a consensus, we will trigger the Reflection mechanism.

\subsection{Reflection} 
As shown in Eq:~\ref{eq:ans}, we have obtained the answer set $S_{ans}$ and reason set $S_{R}$ in the Action process.
Then, each agent is tasked with scoring both its own and other agents' reasoning processes for arriving at the answers formulated as:
\begin{equation}
    Score(A_i) = \sum_{j}^{n}A_j(Q, S_{ans}, S_{R})
\label{eq:judge}
\end{equation}
We can obtain a score set $S_{score}$, as shown in Figure~\ref{fig:method} Reflection stage 1.
If an agent's reasoning is deemed insufficient, it will receive a low score. The lowest score in $S_{score}$ indicates that the agent's performance on this particular long video question is suboptimal. Then this agent will be excluded from further discussion in this question, as shown in Figure~\ref{fig:method} Reflection stage 2. 
Finally, each remaining agent independently summarizes the answers provided by all agents in the previous round, as well as the historical information related to the question. 
Leveraging this historical information, each agent then regenerates the key information required to answer the question. 

\begin{equation}
    K_i = A_i(S_{Score},S_R, S_{ans})
\label{eq:ans}
\end{equation}
This newly generated key information $K_i$ is then utilized to initiate the next round of retrieval, thereby enabling the model to iteratively refine its approach to answering the long video question.

\subsection{Multi-Round Dynamical Collaboration}

In order to obtain a more comprehensive understanding of long video, we introduce a multi-round dynamical collaboration pipeline.
The overall steps of multi-round dynamical collaboration of MLLM agents are shown in Figure~\ref{fig:discuss}.
The upper part of Figure~\ref{fig:discuss} illustrates the whole process of our multi-round discussion. 
In the first round of collaboration, three agents, respectively, perform perception, action, and reflection processes to interact with each other. After that, each subsequent collaboration round involves the cycle of these three processes.
The lower part of Figure~\ref{fig:discuss} presents the specific output results of each agent. In the first and second rounds of discussion, the models did not reach a unified result. However, after multiple rounds of discussion, the models obtained the correct result.

\section{Experiment}
\label{experiment}

\subsection{Benchmark}
We evaluate LVAgent on four long-term video question-answering benchmarks. EgoSchema~\cite{egoschema} consists of over 5,031 human curated question answer pairs derived from Ego4D. EgoSchema requires the correct answer for each question based on a three-minute video clip. 
VideoMME~\cite{videomme} 
includes 900 videos with a total of 254 hours and 2,700 question-answer pairs. The video durations range from 11 seconds to 1 hour, with an average duration of 1024 seconds. 
MLVU~\cite{MLVU} includes 1,730 videos and 2,174 questions in 9 categories, specifically designed for understanding long videos. The videos last between 3 minutes and over 2 hours, with an average length of 930 seconds.  LongVideoBench~\cite{wu2024longvideobench} (LVBench) includes 1,337 QA pairs in val split with their subtitles in various themes. The videos cover four progressive duration groups: 8-15 seconds, 15-60 seconds, 3-10 minutes, and 15-60 minutes, and the average duration is 473 seconds.

\begin{table}[t]

    \begin{minipage}[t]{0.98\linewidth}
        \centering
        \small
        \setlength{\tabcolsep}{3pt} 
        \renewcommand{\arraystretch}{1} 
        \resizebox{1\textwidth}{!}{
            \begin{tabular}{c|cccc}
                \toprule
                \textbf{Model} & \textbf{Frames} & \textbf{Inference Time}  & \textbf{LVBench} & \textbf{VideoMME} \\
                \midrule
                Qwen2-VL\cite{qwen2vl}   & 568 &     90.5s    & 55.6        &  71.2 / 77.8  \\
                
                GPT-4o\cite{gpt4o}      & 384 &     153.6s       & 66.7     &  71.9 / 77.2   \\
 
                Gemini-1.5-Pro\cite{gemini15}   & 568 &    227.2s    & 64.0  &  75.0 / 81.3     \\
                
                \rowcolor{gray!20}
                \textbf{LVAgent}  & \textbf{71.2}   &  \textbf{33.6s} & \textbf{80.0}  & \textbf{ 81.7 / 86.6}\\

                \bottomrule
            \end{tabular}
        }
\vspace{-0.3cm}
        
        \caption{Comparsion on Average Frame Number and Inference Latency on VideoMME.}
        \label{tab:performance}
    \end{minipage}

\vspace{-0.5cm}

\end{table}

\begin{table}[t]

    \begin{minipage}[t]{0.98\linewidth}
        \centering
        \setlength{\tabcolsep}{3pt}
        \renewcommand{\arraystretch}{1}
        \resizebox{1\textwidth}{!}{
            \begin{tabular}{cc|cccc}
                \toprule
                \textbf{Perception} & \textbf{Reflection} & \textbf{Egoschema} & \textbf{LVBench} & \textbf{MLVU} & \textbf{VideoMME} \\
                
                \midrule
                
                \XSolidBrush& \XSolidBrush & 77.9 & 63.9 & 75.7 & 72.1/77.8  \\
                                
                \Checkmark & \XSolidBrush& 80.2 & 69.0 & 79.2  &75.3/81.4 \\

                \XSolidBrush & \Checkmark & 80.4 & 72.2 & 79.3 & 77.6/83.5 \\
                  
                \rowcolor{gray!20}
                \Checkmark & \Checkmark & \textbf{82.9} & \textbf{80.0} & \textbf{83.9} & \textbf{81.7/86.6} \\
                
                \bottomrule
                
            \end{tabular}
        }
        
        \vspace{-0.5em}
        
        \caption{Ablation study on key steps of LVAgent.}
        \label{tab:ablation_com}
    \end{minipage}

\vspace{-2.5em}
    
\end{table}

\subsection{Implication Details} 
The model libraries we construct include Qwen2-VL-7B/72B~\cite{qwen2vl}, LLaVA-Video-72B~\cite{llavavideo}, LongVU-7B~\cite{longvu}, InternVL-2.5-7B/78B~\cite{internvl2.5}, Oryx~\cite{liu2024oryx} and Aria~\cite{aria}. To avoid randomness in the results, when processing, we adjust the temperature coefficient of all MLLMs to 0.01 and uniformly set the maximum number of new tokens to be generated to 168, thus preventing the generation of excessive redundant information.
For the retrieval model, we use ASP-CLIP~\cite{asp} model for better temporal context modeling. We initialize the visual and textual encoders with the pretrained CLIP, following the previous methods~\cite{ref5, ref6, ref7}.
To construct temporal information, we utilize a four-layer temporal transformer featuring 8 attention heads. 
The initial learning rate is configured as 1e-7 for both the text encoder and video encoder, while for other parameters, it is set to 1e-4.
Our ASP-CLIP is optimized using Adam~\cite{ref43}, with a batch size of 64. 
The model is trained for 10 epochs with a cosine learning rate schedule. 
The training process and our experiments on LVAgent are based on Pytorch with 8 A800-80G GPUs.

\subsection{Comparison with SOTA}

As shown in Table~\ref{tab:results}, we compare the methods with closed-source models~\cite{gpt4o, gemini15}, open-source MLLMs~\cite{chen2024sharegpt4video,videochat,internvl2.5,qwen2vl,longva, longvu, llavavideo, aria, liu2024oryx, llavaonevision,videollama2}, and agent-based systems~\cite{videoagentrepeat,videotree, vca,drvideo,videoragvideofeat}. 
Experimental results show that our method is the first agent-based method to achieve over 80\% accuracy across these four benchmarks. 
Especially on LongVideoBench, LVAgent outperforms GPT-4o (SOTA) by 13.3\%. On MLVU, LVAgent surpasses GPT-4o and InternVL-2.5 by 19.3\% and 8.2\%. 

\begin{table}[t]

    \begin{minipage}[t]{0.98\linewidth}
    \small
        \centering
        \renewcommand{\arraystretch}{0.8}
        \resizebox{1\textwidth}{!}{
            \begin{tabular}{c|ccc|cc}
                \toprule
            \begin{tabular}[c]{@{}c@{}}\textbf{Num}\\\textbf{Rounds}  \end{tabular}  &\textbf{LV-72B} & \textbf{Intern-8B} & \textbf{Intern-78B} & \textbf{LVBench} & \textbf{MLVU}\\
                
                \midrule
                
                 \multirow{4}{*}{1} &\Checkmark &  \XSolidBrush      &  \XSolidBrush             &  68.3  &   78.6 \\
                &\XSolidBrush  & \Checkmark&      \XSolidBrush     &  67.8  &   77.8   \\
                & \XSolidBrush     &     \XSolidBrush   &    \Checkmark   &  68.3  &   78.2 \\ 
                
                & \Checkmark     &     \Checkmark   &    \Checkmark   &  69.0   &   79.2 \\ \midrule
                \multirow{4}{*}{2}&\Checkmark  &  \Checkmark  &   \XSolidBrush     &  72.2    &   79.9 \\
                &\XSolidBrush &  \Checkmark  &  \Checkmark        &  75.6     &  80.2  \\
                & \Checkmark & \XSolidBrush   &  \Checkmark       &  77.5   &   80.8 \\ 
                & \Checkmark & \Checkmark    &  \Checkmark       &  77.8   &   81.3 \\ \midrule

                \rowcolor{gray!20}
                3 &\Checkmark & \Checkmark & \Checkmark & \textbf{80.0} & \textbf{83.9} \\
                
                \bottomrule
                
            \end{tabular}
        }
        
        \vspace{-0.75em}
        \caption{Ablation study on different numbers of agents.}
        \label{tab:ablation_agent_num}
    \end{minipage}

\vspace{-1.5em}
    
\end{table}

\paragraph{Efficiency of LVAgent.}  
We evaluate LVAgent's efficiency on the VideoMME benchmark, comparing it with state-of-the-art models in inference time and performance (Table~\ref{tab:performance}). LVAgent processes 71.2 frames and 33.6s of inference time. 
Compared with the current MLLM models that perform best in long video tasks, 
our LVAgent can achieve results that significantly surpass these MLLMs while using fewer frames 
and less inference time.

\subsection{Ablation Study}

\paragraph{Ablation on key steps.}  
We conduct ablation studies to evaluate the contributions of Perception and Reflection processes (Table~\ref{tab:ablation_com}). Results show both components significantly enhance performance on long video understanding tasks. 
Without Perception, agents perform global sampling of 128 frames for fair comparison. 
Using only Reflection or Perception yields improvements, but their combination achieves the best results, attaining 82.9\% on EgoSchema and 80.0\% on LongVideoBench.

\paragraph{Effect of Number of Agents.}  
As shown in Table~\ref{tab:ablation_agent_num}, we analyze the impact of agent combinations on LVBench and MLVU tasks using LLaVA-Video-72B (LV-72B), InternVL-2.5-8B (Intern-8B), and InternVL-2.5-78B (Intern-78B).
Individually, LV-72B, Intern-8B, and Intern-78B achieve 68.3\%, 67.8\%, and 69.0\% on LVBench, respectively. However, combining all three agents yields superior performance (80.0\% on LVBench and 83.9\% on MLVU). 
This improvement stems from the diversity of agents, enabling the model to leverage their complementary strengths for more robust reasoning.

\begin{figure}[t]
    \centering
    \includegraphics[width=0.4\textwidth]{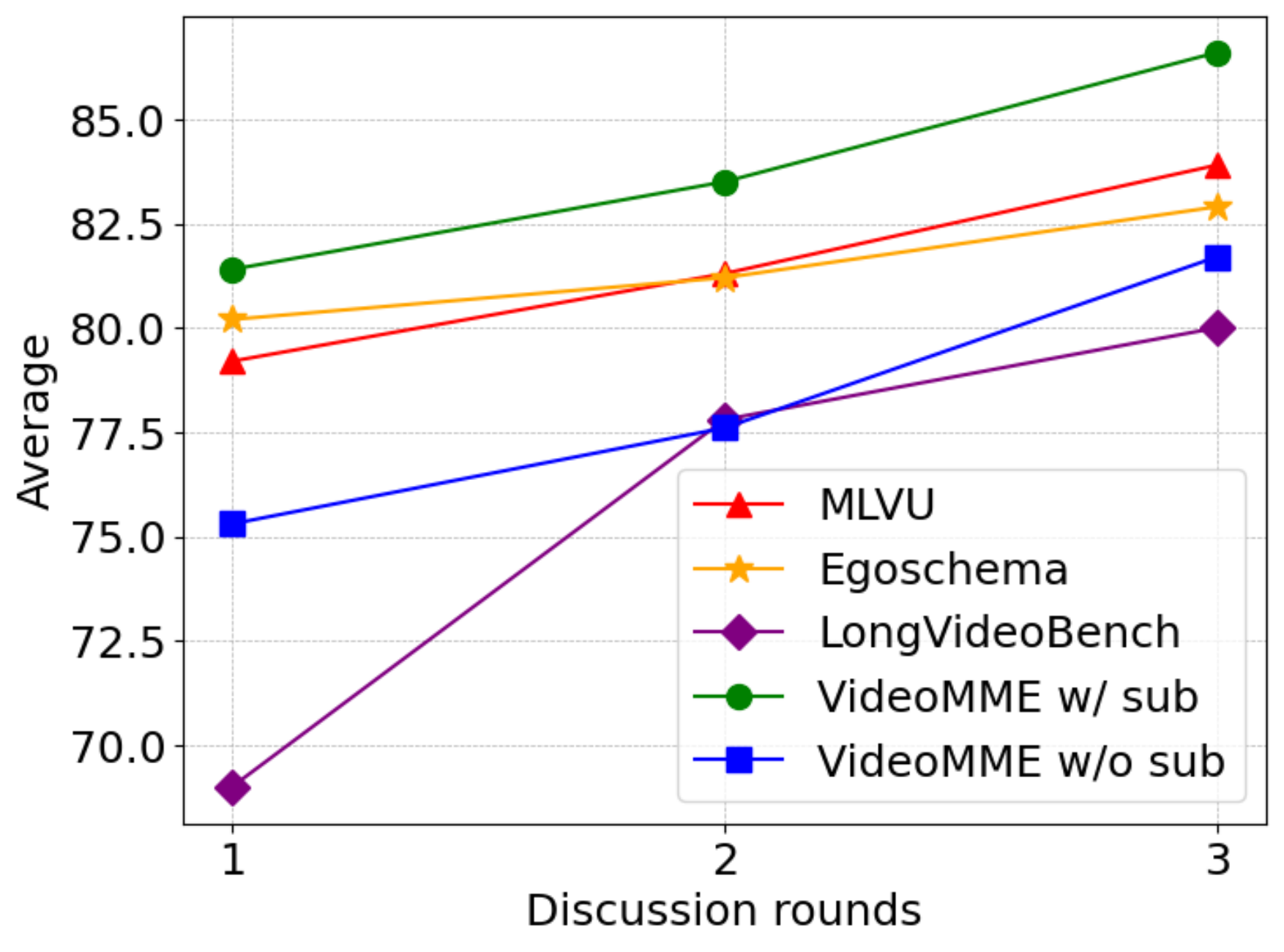}
    \vspace{-0.3cm}
    \caption{
        \textbf{Ablation on Multi-Round Discussion.}
    }
    \label{fig:dis_rouds}
    \vspace{-1.5em}
\end{figure}

\paragraph{Collaboration Rounds of Reflection.}
We investigate the effect of varying the number of maximum reflection rounds one, two, and three on multiple benchmarks, as illustrated in Figure~\ref{fig:dis_rouds}. Increasing the number of rounds consistently enhances performance; for instance, LongVideoBench sees an increase from 69.0\% to 80.0\%. This improvement is attributed to the iterative refinement and consensus among agents, where multiple rounds progressively eliminate noisy knowledge and converge on more accurate answers, significantly boosting performance in complex reasoning.

\paragraph{Effectiveness of ASP-CLIP Finetuning.}
We further explore the impact of different retrieval models used in the Perception process, as shown in Table~\ref{tab:clip}. We show the result of ASP-CLIP pretrained on our collected LongVR dataset. 
The results indicate that ASP-CLIP outperforms the other models.

\paragraph{Ablation on Different Discussion Methods.}
To validate the effectiveness of our multi-round dynamic collaboration, we verify three settings as shown in Table~\ref{tab:fin}.
In the Best score setting, we do not filter out models during each round of the Reflection process. At the third round of discussion, the answer of the model with the highest score is selected as the final answer.
In the decide by agent setting, we do not filter out models either. If no consensus is reached in the final round of discussion, we select the model that performs best in the Selection process and input all the discussion history information, video, question, and options into this model to obtain the final result.
Experiments show that our dynamic collaboration is more effective than other static mechanisms and saves more computational resources.

\paragraph{Ablation on the Retrieval Threshold.}
To determine the optimal retrieval threshold, we conduct an ablation study, as shown in Table~\ref{tab:threshold}. The results indicate that setting the retrieval threshold to 0.8 yields the best performance across all benchmarks. This is likely because a higher threshold ensures that only the most critical video frames are selected, leading to more accurate and efficient retrieval.

\begin{table}[t]

    \begin{minipage}[t]{0.98\linewidth}
        \vspace{0pt}
        \centering
        \small
        
        \resizebox{\textwidth}{!}{
            \begin{tabular}{c|cccc}
                \toprule
                \begin{tabular}[c]{@{}c@{}}\textbf{Retrieval}\\\textbf{Model}  \end{tabular}  & \textbf{Egoschema } & \textbf{LVBench} & \textbf{MLVU} & \textbf{VideoMME} \\
                \midrule
               CLIP & 80.9 & 74.7 & 81.3 &  78.7/84.0 \\
               
                LongCLIP & 81.2 & 75.6 & 82.2 &  79.6/85.4 \\
                \rowcolor{gray!20}
                \textbf{ASP-CLIP} & \textbf{82.9} & \textbf{80.0} & \textbf{83.9} & \textbf{81.7/86.6} \\
                \bottomrule
                
            \end{tabular}
        }
        
        \vspace{-5pt}
        
        \caption{Ablation study on the retrieval model in Perception.}
        \label{tab:clip}
    \end{minipage}

\vspace{-0.3cm}
\end{table}

\begin{table}[t]

    \begin{minipage}[t]{0.98\linewidth}
        \centering
        \setlength{\tabcolsep}{3pt}
        \renewcommand{\arraystretch}{1.2}
        \resizebox{1\textwidth}{!}{
            \begin{tabular}{c|cccc}
                \toprule
                \textbf{Discussion Methods} & \textbf{Egoschema } & \textbf{LVBench} & \textbf{MLVU} & \textbf{VideoMME} \\
                \midrule
                Best Score     &  81.3    & 76.9  &   80.2  & 78.8 / 82.1   \\
                Decide by Agent &  81.2    &  77.0     &   79.9  & 79.1 / 82.2  \\
                \rowcolor{gray!20}
                
                \textbf{Dynamic Collaboration} & \textbf{82.9} & \textbf{80.0} & \textbf{83.9} & \textbf{81.7 / 86.6} \\
                \bottomrule
                
            \end{tabular}
        }
        
        \vspace{-0.5em}
        \caption{Ablation study on different discussion methods.}
        
        \label{tab:fin}
    \end{minipage}
    \vspace{-1em}

\end{table}

\begin{table}[t]
    \begin{minipage}[t]{0.98\linewidth}
        \renewcommand{\arraystretch}{1.1}
        \centering
        \resizebox{\textwidth}{!}{
            \begin{tabular}{c|cccc}
                \toprule
                \begin{tabular}[c]{@{}c@{}}\textbf{Retrieval}\\\textbf{Threshold}  \end{tabular} & \textbf{EgoSchema } & \textbf{LongVideoBench} & \textbf{MLVU} & \textbf{VideoMME} \\
                \midrule
                0.4 & 81.7 & 78.6 & 82.2 &  79.6/82.2 \\
                0.6 & 82.4 & 78.9 & 82.9 &  80.1/83.1 \\
                \rowcolor{gray!20}
               \textbf{ 0.8 }&\textbf{ 82.9} &\textbf{ 80.0 }& \textbf{83.9 }&  \textbf{81.7/86.6 }\\
                \bottomrule
                
            \end{tabular}
        }
        
        \vspace{-0.6em}
        
        \caption{
            Ablation study on the retrieval threshold in Perception.
        }
        \label{tab:threshold}
    \end{minipage}
\vspace{-1.5em}
\end{table}

\section{Conclusion}
\label{conclusion}

In this paper, we introduce LVAgent, a multi-agent collaborative framework for long video understanding, addressing the limitations of existing single-model methods that suffer from insufficient reasoning capabilities. 
LVAgent enables multi-round dynamic collaboration among MLLM agents through four key steps: \textit{Selection} of optimal agent teams, \textit{Perception} with efficient temporal retrieval, \textit{Action} for question answering and reasoning exchange, and \textit{Reflection} for iterative refinement. 
Experiments on VideoMME, EgoSchema, MLVU, and LongVideoBench demonstrate LVAgent's superiority, achieving over 80\% accuracy across tasks and up to 13.3\% improvement on LongVideoBench. 
LVAgent is the first multi-round dynamic multi-agent collaboration pipeline that outperforms all closed-source models (including GPT-4o) and open-source models (including InternVL-2.5 and Qwen2-VL) in the long video understanding tasks.
These results highlight the benefits of our multi-agent approach in long video understanding tasks.

\section{Acknowledgement}
This work was supported by the National Key R\&D Program of China (NO.2022ZD0160505).
We sincerely express the gratitude for Zijun Liu on valuable discussions about multi-round dynamical collaboration of MLLM agents.

{
    \small
    \bibliographystyle{ieeenat_fullname}
    \bibliography{main}
}

\newpage
\clearpage
\appendix
\clearpage

\section{Details of LongVR Dataset.}
\label{sec:videoclip_dataset}

We present a detailed breakdown of the data distribution and key statistics for LongVR dataset in Table~\ref{tab:videoclip_dataset}. Specifically, ActivityNet-Caption contains 37,421 instances with an average caption length of 15.80 words and an average video duration of 171.44 seconds. OpenVid-1M comprises 30,000 instances, featuring longer captions with an average length of 143.16 words and shorter videos averaging 9.07 seconds. ViTT has 5,086 instances, with captions averaging 24.98 words and videos lasting 305.08 seconds on average. MovieChat-Caption includes 808 instances, having the longest captions at an average of 143.47 words, and videos with an average duration of 457.65 seconds. Youcook2 consists of 8,700 instances, with captions averaging 80.90 words and videos lasting 383.19 seconds. Collectively, the LongVR dataset encompasses a total of 82,015 instances, with an overall average caption length of 71.12 words and an average video duration of 145.62 seconds. This diverse and extensive dataset enables robust training of the ASP-CLIP model, capturing a wide range of video and caption characteristics.
\begin{table}[htbp]
\centering
\small 
\setlength{\tabcolsep}{3pt} 
\begin{tabular}{crrr}
\toprule
\textbf{Dataset} & \makecell{\textbf{Instance} \\ \textbf{Num}} & \makecell{\textbf{Avg Caption} \\ \textbf{Length}} & \makecell{\textbf{Avg Video} \\ \textbf{Length}} \\
\midrule
ActivityNet-Caption  & 37,421 & 15.80 & 171.44\\
OpenVid-1M  & 30,000 & 143.16 & 9.07\\
ViTT & 5,086  & 24.98 & 305.08 \\
MovieChat-Caption & 808 & 143.47 & 457.65\\
Youcook2  & 8,700 &80.90 & 383.19\\
\midrule
\textbf{Total} & \textbf{82,015}  &\textbf{71.12} & \textbf{145.62} \\
\bottomrule
\end{tabular}
\caption{Instance numbers of different datasets for training the ASP-CLIP model.}
\label{tab:videoclip_dataset}
\vspace{-1em}
\end{table}

\section{More Experiment Details and Results.}

\subsection{Model Selection for Different Tasks}

For VideoMME, we employ InternVL2.5 (78B \& 8B) and Qwen2VL (72B). For EgoSchema, we use Qwen2VL (72B \& 7B) and LLaVAVideo (72B). For MLVU and LongVideoBench, we utilize LLaVAVideo (72B) and InternVL2.5 (78B \& 8B). Our selection mechanism chooses different agent groups based on pseudo labels and dataset-specific features.

\subsection{Efficiency Analysis}

We evaluate a total of 13,943 videos across the datasets LongVideoBench, MLVU, VideoMME (with and without subtitles), and EgoSchema. For comparison, Qwen2VL (72B) requires 350.5 hours in total, averaging 90.5 seconds per video. We break down the time consumption of our approach into the following steps:
Fine-tuning, conducted with 95M parameters, takes 3 hours.
Preselection on 750 videos consumes 6.58 hours, averaging 31.6 seconds per video.
The evaluation phase takes 130.13 hours, with an average of 33.6 seconds per video.
In total, the evaluation process (including preselection and evaluation) requires 139.71 hours, averaging 36.07 seconds per video—this is 2.5× faster than Qwen2VL.

\subsection{The Result of Using More Than Three Agents}

The performance of the 4-agent setup (EgoSchema: 83.0, LongVideoBench: 80.1, MLVU: 84.1, VideoMME: 81.8 / 86.5) is slightly superior to that of the 3-agent configuration, but it incurs additional time cost—taking 38.7 seconds per video, which is 5.1 seconds longer than the 3-agent setup. Considering the trade-off between accuracy and efficiency, we opt for the 3-agent configuration. Notably, LVAgent can be easily extended to incorporate more agents if needed.

\section{Prompts of LVAgent.}
\label{sec:prompts}

In this section, we present the comprehensive set of prompts utilized by LVAgent across various stages of its operation, as shown in Table~\ref{tab:all-prompts}. These prompts are meticulously designed to guide LVAgent in effectively performing its tasks, ensuring seamless interactions and accurate responses.

\clearpage
\begin{onecolumn}
\captionsetup{width=\textwidth}
\begin{longtable}{p{4.8cm}|p{10cm}}
\toprule
\textbf{Prompt} & \textbf{Content} \\ \midrule
\endhead
Prompt to decide whether \newline to watch the whole video & 
You are given a single-choice question, options, subtitles, and some frames of the long video. You should not only look at the textual information but also consider the input visual information, taking everything into account. If you can answer the question accurately and comprehensively based on the existing information, especially the visual information, and further watching the entire video will not significantly improve the quality of the answer, then you don't need to watch the entire video and can answer 'No.'. However, if the existing information is not sufficient to fully answer the question, and watching the entire video may obtain information crucial for answering the question, please reply 'Yes' \newline
The frame tokens: \{\verb|Frame tokens|\}
\newline \{\verb|Question|\} \newline \{\verb|Options|\} \newline \{\verb|Subtitiles|\}(If have)\newline Output:\{\verb|Yes/No|\} 

 \\ \midrule
 Prompt for generate the key information & 

 Given four randomly sampled frames from a long video, subtitles, a question, and multiple-choice options, identify the key information needed to answer the question. Focus on visual cues, context, and temporal relationships within the frames. Limit your response to 50 words. \newline
The frame tokens: \{\verb|Frame tokens|\}
\newline \{\verb|Question|\} \newline \{\verb|Options|\} \newline \{\verb|Subtitles|\}(If have) \newline
  \\ \midrule

Prompt for generating the answer & 
Select the best answer to the following multiple-choice question based on the video and the subtitles. Respond with only the letter (A, B, C, or D) of the correct option. \newline
The frame tokens: \{\verb|Frame tokens|\}
\newline \{\verb|Question|\} \newline \{\verb|Options|\} \newline \{\verb|Subtitles|\} (If have)\newline
The best answer is:
  \\ \midrule
Prompt for generating the reason & Given the video frames you've seen, and the question along with your answer, deeply analyze the logical steps and evidence from the frames that led you to provide this particular answer. \newline The Question is:  \{\verb|Question|\}\newline The predict answer is \{\verb|Predict answer|\} \\ \midrule
Prompt for generating scores for agents & 
 
You are given the answers and the reasoning for judgment from this model and two other models.\newline
The question is: \{\verb|Question|\} \newline
The answer of \{\verb|Agent 1|\} is \{\verb|Agent 1's Answer|\} \newline the reason is \{\verb|Agent 1's Reason|\} \newline
The answer of \{\verb|Agent 2|\} is \{\verb|Agent 2's Answer|\} \newline the reason is \{\verb|Agent 2's Reason|\} \newline
The answer of \{\verb|Agent 3|\} is \{\verb|Agent 3's Answer|\} \newline the reason is \{\verb|Agent 3's Reason|\} \newline
Please score the performance of these three agents based on their reasoning. The score ranges from 1 to 10. \newline
Please strictly follow the answer format\! The answer format is: \newline
\{\verb|Agent 1's Score|\}: 1-10 \newline
\{\verb|Agent 2's Score|\}: 1-10 \newline
\{\verb|Agent 3's Score|\}: 1-10 \newline The reason is: \{\verb|Reason|\} \\ \midrule
Prompt for history information summarization & 
Agent 1's answer is: \{\verb|Agent 1's Answer|\}.\newline
Reason: \{\verb|Agent 1's Reason|\}.\newline
The final score is : \{\verb|Agent 1's Score|\}.\newline
Agent 2's answer is: \{\verb|Agent 2's Answer|\}.\newline
Reason: \{\verb|Agent 2's Reason|\}.\newline
The final score is : \{\verb|Agent 2's Score|\}.\newline
Removed \{\verb|Agent 3|\} Answer\newline Answer: \{\verb|Agent 3's Answer|\}\newline Reason \{\verb|Agent 2's Reason|\}\newline However, this reason was deemed unconvincing, so this answer was removed from the discussion. \\ 
\bottomrule

\caption{The prompting templates used in different key steps of LVAgent.}
\label{tab:all-prompts} 
\end{longtable}
\end{onecolumn}

\end{document}